\documentclass[letterpaper, 10 pt, conference]{ieeeconf}
\IEEEoverridecommandlockouts                
\usepackage{times}
\usepackage{setspace}
\usepackage{url}
\spacing{1}
\usepackage[utf8]{inputenc}
\usepackage[T1]{fontenc}
\usepackage{graphicx}		
\usepackage{wrapfig}
\usepackage[format=plain,font=footnotesize,labelfont=bf,labelsep=period]{caption}
\usepackage{sidecap} 
\usepackage{subfig}
\usepackage[export]{adjustbox}
\usepackage[font=small]{caption}
\usepackage{float}
\usepackage{dblfloatfix}

\usepackage{amsmath} 
\usepackage{amssymb}  
\usepackage{amsthm}
\usepackage{mathtools}
\usepackage[normalem]{ulem}
\usepackage{paralist}	
\usepackage[space]{grffile} 
\usepackage{color}
\usepackage{bbm}
\usepackage{placeins} 
\usepackage{array}
\usepackage{siunitx}
\usepackage{hyperref}

\theoremstyle{definition}
\newtheorem{definition}{Definition}
\theoremstyle{remark}

\theoremstyle{definition}

\theoremstyle{definition}

\theoremstyle{definition}

\newcommand{\R}{\mathbb{R}}
\newcommand{\C}{\mathcal{C}}

\newcommand{\K}{\mathcal{K}}

\newcommand{\gap}{\vspace{.1cm}}

\newcommand{\newsec}[1]{\gap {\bf \noindent #1 }}

\definecolor{darkblue}{RGB}{0,0,102}
\definecolor{lightblue}{RGB}{77,77,148}

\definecolor{gold}{RGB}{234, 170, 0}
\definecolor{metallic_gold}{RGB}{139, 111, 78}

\renewcommand{\cal}[1]{\mathcal{ #1 }}
\newcommand{\mb}[1]{\mathbf{ #1 }}
\newcommand{\bs}[1]{\boldsymbol{ #1 }}

\newcommand{\derp}[2]{\frac{\partial #1 }{\partial #2 }}

\DeclareMathOperator*{\argmin}{argmin}

\newcommand{\new}[1]{{#1}}

\renewcommand{\baselinestretch}{.985}
\allowdisplaybreaks

\newcommand{\HRule}[1][\medskipamount]{\par
  \vspace*{\dimexpr-\parskip-\baselineskip+#1}
  \noindent\rule{\columnwidth}{0.5pt}\par
  \vspace*{\dimexpr-\parskip-0.5\baselineskip+#1}}
\newcommand{\boxedequation}[2]{
\begin{samepage}
\HRule[8pt]
\vspace{-4pt}
\newsec{\textbf{{#1:}}} \nopagebreak
#2
\HRule[8pt]
\end{samepage}
}

\newcommand{\Lim}[1]{\raisebox{0.5ex}{\scalebox{0.8}{$\displaystyle \lim_{#1}\;$}}}
\newcommand*\dif{\mathop{}\!\mathrm{d}}

\newcommand{\vth}{\mbox{\boldmath $\theta$}}

\newcommand{\vlambda}{\mbox{\boldmath $\lambda$}}

\newcommand{\vtau}{\mbox{\boldmath $\tau$}}

\newcommand{\vom}{\mbox{\boldmath $\omega$}}

\newcommand{\vb}{\mathbf b}

\newcommand{\vg}{\mathbf g}
\newcommand{\vh}{\mathbf h}

\newcommand{\vn}{\mathbf n}

\newcommand{\vp}{\mathbf p}
\newcommand{\vq}{\mathbf q}
\newcommand{\vr}{\mathbf r}

\newcommand{\vu}{\mathbf u}
\newcommand{\vv}{\mathbf v}

\newcommand{\vx}{\mathbf x}

\newcommand{\vA}{\mathbf A}

\newcommand{\vD}{\mathbf D}

\newcommand{\vF}{\mathbf F}

\newcommand{\vI}{\mathbf I}

\newcommand{\vP}{\mathbf P}

\newcommand{\vR}{\mathbf R}

\newcommand{\vT}{\mathbf T}

\title{\LARGE \textbf{Multi-Layered Safety for Legged Robots via \\ Control Barrier Functions and Model Predictive Control}}
\author{Ruben Grandia, Andrew J. Taylor, Aaron D. Ames, Marco Hutter
\thanks{R. Grandia and M. Hutter are supported via the European Union’s Horizon 2020 research and innovation programme under grant agreement No 780883 and by the Swiss National Science Foundation (SNSF) as part of project No.188596. A. Taylor, and A. Ames are supported via DARPA award HR00111890035, and NSF awards 1923239 and 1924526.}
\thanks{R. Grandia and M. Hutter are with the Department of Mechanical and Process Engineering, ETH Z\"urich, 8092 Z\"urich, Switzerland {\tt\small \{rgrandia,mahutter\}@ethz.ch}. A. Taylor and A. Ames are with the Department of Computing and Mathematical Sciences, California Institute of Technology, Pasadena, CA 91125, USA {\tt\small \{ajtaylor,ames\}@caltech.edu}.
}
}
\begin{document}
\nocite{video}

\maketitle
\thispagestyle{empty}
\pagestyle{empty}

\begin{abstract}
The problem of dynamic locomotion over rough terrain requires both accurate foot placement together with an emphasis on dynamic stability. Existing approaches to this problem prioritize immediate safe foot placement over longer term dynamic stability considerations, or relegate the coordination of foot placement and dynamic stability to heuristic methods. We propose a multi-layered locomotion framework that unifies Control Barrier Functions (CBFs) with Model Predictive Control (MPC) to simultaneously achieve safe foot placement and dynamic stability. Our approach incorporates CBF based safety constraints both in a low frequency kino-dynamic MPC formulation and a high frequency inverse dynamics tracking controller. This ensures that safety-critical execution is considered when optimizing locomotion over a longer horizon. We validate the proposed method in a 3D stepping-stone scenario in simulation and experimentally on the ANYmal quadruped platform.
\end{abstract}

\section{Introduction}
\label{sec:intro}
A key motivation behind the development of legged robots is their ability to overcome complex terrain. Because legged locomotion only requires discrete footholds, obstacle such as steps, gaps, and stairs can be traversed, making legged robots a compelling alternative to wheeled systems. When a statically stable motion pattern is considered, several mature strategies for rough terrain locomotion have been proposed and successfully demonstrated on hardware for  bipedal~\cite{griffin2019footstep}, quadrupedal~\cite{fankhauser2018robust,mastalli2018motion}, and hexapedal~\cite{Belter2016} robots. However, inspired by the fast and dynamic motions seen in nature, the use of dynamic gaits---a gait where individual contact phases are statically unstable---is still an active area of research. 

\begin{figure}[t]
    \centering
    \includegraphics[width=\columnwidth]{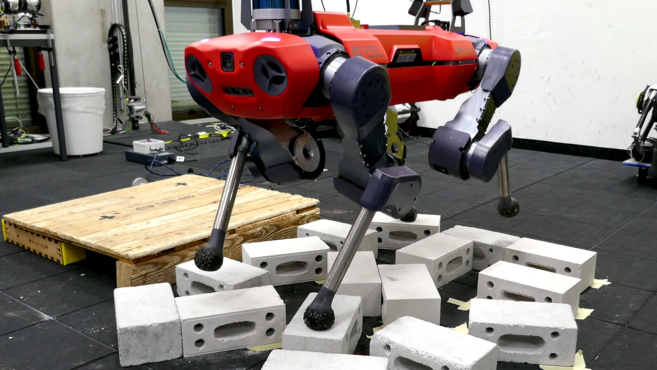}
    \caption{ANYmal~\cite{Hutter2016ANYmal} performing a trotting gait on stepping-stones.}
    \label{fig:anymal}
    \vspace{-5mm}
\end{figure}

The challenge in dynamic locomotion lies in the fact that foothold locations are not only constrained by the terrain, but also affect the dynamic stability of the resulting contact configuration. Additionally, as the speed of the motions increases, the inertial and nonlinear effects described by the full rigid body dynamics of the system become more relevant. There is therefore a need for methods that can guarantee a safe foot placement while simultaneously considering the future impact on the dynamic stability of the system. A classical locomotion challenge that demands safe foot placement and dynamic stabilization is the “stepping-stones” scenario, see Figure \ref{fig:anymal}, where viable foothold locations are discontinuous and sparsely available. We propose to combine the safety guarantees endowed by Control Barrier Functions (CBFs) with the longer horizon considered in Model Predictive Control (MPC) to guarantee safe foot placement while achieving dynamic locomotion and high tracking performance.

\subsection{Related work}

Control Barrier Functions \new{\cite{ames2014control}} are a tool for synthesizing controllers that ensure safety of nonlinear systems \new{ \cite{jankovic2018robust,ames2019control}}. Moreover, CBFs have been used in the stepping-stones problem via a quadratic programming (QP) based tracking controller~\cite{nguyen2015optimal,nguyen20163d}. An offline optimized walking trajectory, or a library thereof~\cite{nguyen2020dynamic}, is tracked and locally modified to satisfy CBF safety constraints. While promising in simulation, we are not aware of the successful transfer of a CBF based stepping controller to hardware, despite extensions that add robustness~\cite{nguyen2016optimal}, or a learning based model error correction~\cite{choi2020reinforcement}. Indeed, in~\cite{nguyen2018dynamic}, the stepping-stones problem is demonstrated experimentally by increasing the look-ahead horizon of the gait library and through subsequent gait interpolation rather than a CBF based method. We hypothesize that it is exactly this reasoning over a longer horizon that is missing with the CBF-QP control formulation.

In contrast, Model Predictive Control has become a central method for the online synthesis and control of dynamic systems over a given time horizon \cite{rawlings2017model}. In the context of the stepping-stones problem, a distinction can be made between MPC based approaches where the footholds locations are determined separately from the torso motion optimization \cite{jenelten2020perceptive,villarreal2020mpc,Kim2020vision}, and MPC based approaches where the foothold location and torso motions are jointly optimized. The benefit of jointly optimizing torso and leg motions has been demonstrated in the field of trajectory optimization~\cite{Dai2014,Winkler2018}. Following this idea, real-time capable methods have been proposed with the specification of leg motions made at the position~\cite{Bledt2017}, velocity~\cite{Farshidian2017realtime}, or acceleration level~\cite{Neunert2018}. One challenge of this approach is its computational costs, which can be resolved by coupling a low-frequency MPC controller with a high-frequency tracking controller \cite{grandia2019feedback}.

\begin{figure}[t!]
    \centering
    \includegraphics[width=0.9\columnwidth]{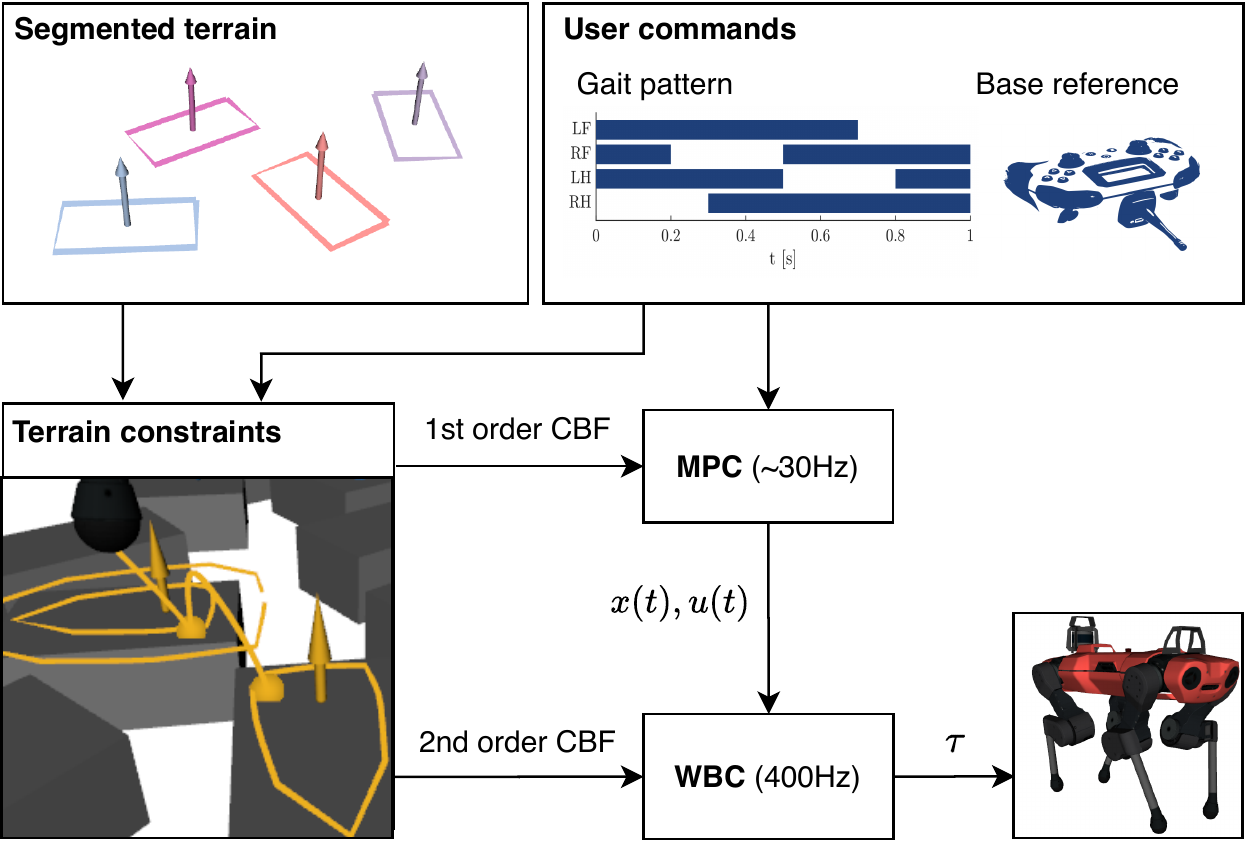}
    \caption{Overview of the proposed multi-layed control setup showing both the MPC and WBC layer receiving terrain CBF constraints.}
    \label{fig:control_overview}
    \vspace{-4mm}
\end{figure}

\subsection{Contribution}

In this work, we build upon a kino-dynamic MPC formulation~\cite{Farshidian2017realtime} where joint velocities and contact forces are decision variables in a low frequency MPC controller. This allows direct integration of CBF safety constraints into the MPC formulation similar to \cite{grandia2020nonlinear}. \new{By jointly optimizing torso and leg motions our method avoids the heuristic coordination that is needed when foot placement and torso motion are delegated to separate controllers}. A higher rate tracking controller is implemented that fuses inverse dynamics with the CBF safety constraints to offer guarantees of safety with the whole-body dynamics in consideration.  \new{In the context of collision avoidance, CBFs can be thought of (and have shown to be) a generalization  of artificial potential fields used in inverse dynamics methods~\cite{Khatib1985}\cite{singletary2020comparative}.} Finally, we note that the combination of discrete time CBFs with MPC has been considered in \cite{zeng2020safety}, but it did not consider a multi-layered approach nor provided experimental results. 

The main contributions of this work are two-fold. First, we propose a multi-layered control approach that combines CBFs with MPC (see Figure \ref{fig:control_overview}). This framework allows CBF safety constraints on the position coordinates of robotic systems to be incorporated in a low frequency MPC controller determining desired velocities as well as in a high frequency tracking controller that incorporates the dynamics of the system. Compared to standard CBF approaches, this adds a horizon when determining safe control inputs. Compared to MPC approaches, the safety critical constraint is enforced at a higher rate, and incorporates a higher fidelity whole-body dynamics model. The second contribution is, to the best of the author's knowledge, the first successful experimental demonstration of CBFs, not only as an approach to the stepping-stones problem, but on a legged robot.


\section{Background}
\label{sec:background}
This section provides a review of Control Barrier Functions (CBFs) and Nonlinear Model Predictive Control. 

Consider the nonlinear control affine system given by:
\begin{equation}
    \label{eqn:eom}
    \dot{\mb{x}} = \mb{f}(\mb{x})+\mb{g}(\mb{x})\mb{u},
\end{equation}
where $\mb{x}\in\R^n$, $\mb{u}\in\R^m$. $\mb{f}:\R^n\to\R^n$ and $\mb{g}:\R^n\to\R^{n\times m}$ are locally Lipschitz continuous on $\R^n$. Given a Lipschitz continuous state-feedback controller $\mb{k}:\R^n\times\R_+\to\R^m$, the closed-loop system dynamics are:
\begin{equation}
    \label{eqn:cloop}
    \dot{\mb{x}} = \mb{f}_{\textrm{cl}}(\mb{x},t) \triangleq  \mb{f}(\mb{x})+\mb{g}(\mb{x})\mb{k}(\mb{x},t).
\end{equation}
The assumption on local Lipschitz continuity of $\mb{f}$, \new{$\mb{g}$} and $\mb{k}$ implies that $\mb{f}_\textrm{cl}$ is locally Lipschitz continuous. Thus for any initial condition $\mb{x}_0 := \mb{x}(0) \in \R^n$ there exists a maximum time interval $I(\mb{x}_0) = [0, t_{\textrm{max}})$ such that $\mb{x}(t)$ is the unique solution to \eqref{eqn:cloop} on $I(\mb{x}_0)$ \cite{perko2013differential}.

\subsection{Control Barrier Functions}

The notion of safety that we consider in this paper is formalized by specifying a \textit{safe set} in the state space that the system must remain in. In particular, consider a time-varying set $\C_t\subset \R^n$ defined as the 0-superlevel set of a continuously differentiable function $h:\R^n\times\R_+ \to \R$, yielding:
\begin{align}
    \C_t &\triangleq \left\{\mb{x} \in \R^n : h(\mb{x},t) \geq 0\right\}, \label{eqn:safeset}
\end{align}
We refer to $\C_t$ as the \textit{safe set}. This construction motivates the following definitions of forward invariant and safety:

\begin{definition}[\textit{Forward Invariant \& Safety}]
A time-varying set $\C_t\subset\R^n$ is \textit{forward invariant} if for every $\mb{x}_0\in\C_0$, the solution $\mb{x}(t)$ to \eqref{eqn:cloop} satisfies $\mb{x}(t) \in \C_t$ for all $t \in I(\mb{x}_0)$. The system \eqref{eqn:cloop} is \textit{safe} on the set $\C_t$ if the set $\C_t$ is forward invariant.
\end{definition}


\noindent Certifying the safety of the closed-loop system \eqref{eqn:cloop} with respect to a set $\C_t$ may be impossible if the controller $\mb{k}$ was not chosen to enforce the safety of $\C_t$. Control Barrier Functions can serve as a synthesis tool for attaining the forward invariance, and thus the safety of a set. Before defining CBFs, we note a continuous function $\alpha:(-\infty,\infty)\to\R$, is said to belong to \textit{extended class $\cal{K}_\infty$ ($\alpha\in\cal{K}_{\infty,e}$)} if $\alpha$ is strictly monotonically increasing, $\alpha(0)=0$, and if $\Lim{r\to\infty}\alpha(r)=\infty$, and $\Lim{r\to-\infty}\alpha(r)=-\infty$.
\begin{definition}[\textit{Control Barrier Function (CBF)}, \cite{ames2017control}]
Let $\C_t\subset\R^n$ be the time-varying 0-superlevel set of a continuously differentiable function $h:\R^n\times\R_+\to\R$ with $0$ a regular value. The function $h$ is a time-varying \textit{Control Barrier Function} (CBF) for \eqref{eqn:eom} on $\C_t$ if there exists $\alpha\in\K_{\infty,e}$ such that for all $\mb{x}\in\R^n$ and $t\in\R_+$:
\begin{align}
\label{eqn:cbf}
     \sup_{\mb{u}\in\R^m} \dot{h}(\mb{x},t,\mb{u}) \triangleq &\derp{h}{\mb{x}}(\mb{x},t)\left(\mb{f}(\mb{x})+\mb{g}(\mb{x})\mb{u}\right)\nonumber\\&+\derp{h}{t}(\mb{x},t)  \geq-\alpha(h(\mb{x},t)).
\end{align}
\end{definition}
\noindent \new{Controllers that take inputs satisfying \eqref{eqn:cbf} ensure the safety of the closed-loop system \eqref{eqn:cloop} \cite{ames2014control}}. 

Given a nominal (but not necessarily safe) locally Lipschitz continuous controller $\mb{k}_d:\R^n\times\R_+\to\R^m$, a \new{possible} controller taking values \new{satisfying \eqref{eqn:cbf}} is the safety-critical CBF-QP:
\begin{align}
\label{eqn:CBF-QP}
\tag{CBF-QP}
\mb{k}(\mb{x},t) =  \,\,\underset{\mb{u} \in \R^m}{\argmin}  &  \quad \frac{1}{2} \| \mb{u} -\mb{k}_d(\mb{x},t)\|_2^2  \\
\mathrm{s.t.} \quad & \quad \dot{h}(\mb{x},t,\mb{u})
\geq -\alpha(h(\mb{x},t)). \nonumber
\end{align}

\subsection{Nonlinear Model Predictive Control}
We consider the following nonlinear optimal control problem with cost functional 
\begin{equation}
\min_{u(\cdot)}   \Phi(\vx(T)) + \int_{0}^{T} L(\vx(t),\vu(t), t) \, \dif t, 
 \label{eq:mpc_cost}
\end{equation}
where $\vx(t)$ is the state, $\vu(t)$ is the input at time $t$, $L(\cdot)$ is an intermediate cost, and $\Phi(\cdot)$ is the cost at the terminal state $\vx(T)$.
The goal is to find a continuous control signal $\mb{u}:I(\mb{x}_0)\to\R^m$ that minimizes this cost subject to the system dynamics, initial condition, and general constraints:
\begin{align}
& \dot{\vx} =  f(\vx, \vu, t),  \label{eq:mpc_dynamics} \qquad \vx(0) = \vx_0, \\
& \vg(\vx,\vu, t) =  0, \qquad
 \vh(\vx,\vu, t) \geq  0 \label{eq:mpc_inequality}.
\end{align}

\noindent Various methods exist to solve this problem~\cite{rawlings2017model}, and a detailed discussion is beyond the scope of this paper. 
In this work we use the Sequential Linear Quadratic (SLQ) method, which is a Differential Dynamic Programming (DDP) based algorithm for continuous-time systems. In particular, the method in~\cite{grandia2019feedback} is being used which extends the (SLQ) formulation of~\cite{Farshidian2017} for use with inequality constraints.

\section{Multi-Layered Control Formulation}
\label{sec:formulation}
In this section we present a multi-layered control formulation that unifies CBFs with MPC to achieve safety and longer horizon optimality for a general robotic system. Consider a robotic system with generalized coordinates $\mb{q}\in\R^d$ and coordinate rates $\dot{\mb{q}}\in\R^d$ with dynamics given by: 
\begin{equation}
    \mb{D}(\mb{q})\ddot{\mb{q}}+\mb{C}(\mb{q},\dot{\mb{q}})\dot{\mb{q}}+\mb{G}(\mb{q}) = \mb{B}(\mb{q})\bs{\tau},
    \label{eq:full_rigid_body_dynamics}
\end{equation}
with inertia matrix $\mb{D}$, centrifugal and Coriolis terms $\mb{C}$, gravitational forces $\mb{G}$, actuation matrix $\mb{B}$, and torques $\bs{\tau}\in\R^m$. Consider a continuously differentiable function $h:\R^d\times\R_+\to\R$ that determines a time-varying safe set for the position coordinates of the robot, with a time derivative given by:
\begin{equation*}
    \dot{h}(\mb{q},\dot{\mb{q}},t) = \derp{h}{\mb{q}}(\mb{q},t)\dot{\mb{q}}+\derp{h}{t}(\mb{q},t).
\end{equation*}
The torques $\bs{\tau}$ do not appear in this time derivative, making it impossible to choose inputs that ensure the barrier constraint:
\begin{equation}
    \label{eqn:reldeg1}
    \dot{h}(\mb{q},\dot{\mb{q}},t) \geq - \alpha_1(h(\mb{q},t)),
\end{equation}
is met for some $\alpha_1\in\K_{\infty,e}$. This challenge is often resolved through the notion of \textit{exponential} CBFs \cite{nguyen2016exponential}, in which an auxiliary function $h_e:\R^d\times\R^d\times\R_+\to\R$ is defined as:
\begin{equation}
    h_e(\mb{q},\dot{\mb{q}},t) = \dot{h}(\mb{q},\dot{\mb{q}},t) + \alpha_1(h(\mb{q},t)),
\end{equation}
\begin{equation*}
    \dot{h}_e(\mb{q},\dot{\mb{q}},t,\bs{\tau}) = \derp{h_e}{\mb{q}}(\mb{q},\dot{\mb{q}},t)\dot{\mb{q}} + \derp{h_e}{\dot{\mb{q}}}(\mb{q},\dot{\mb{q}},t)\ddot{\mb{q}}+\derp{h_e}{t}(\mb{q},\dot{\mb{q}},t).
\end{equation*}
As $\ddot{\mb{q}}$ appears in affine relation to $\bs{\tau}$ in~\eqref{eq:full_rigid_body_dynamics}, $h_e$ can serve as a CBF for the set $\C_{t,e} \triangleq \left\{(\mb{q},\dot{\mb{q}}) \in \R^{2d} : h_e(\mb{q},\dot{\mb{q}},t) \geq 0\right\}$ by enforcing:
\begin{equation}
\label{eqn:reldeg2}
    \dot{h}_e(\mb{q},\dot{\mb{q}},t,\bs{\tau}) \geq - \alpha_2(h_e(\mb{q},\dot{\mb{q}},t)),
\end{equation}
for some $\alpha_2\in\K_{\infty,e}$. Enforcing the forward invariance of this set implies the desired safety constraint \eqref{eqn:reldeg1} is met, implying the forward invariance of the set $\C_t\cap\C_{t,e}$. Thus the constraint on the position coordinates of the robot are met. 

Typical approaches using exponential CBFs only enforce the final constraint \eqref{eqn:reldeg2}, often in a \ref{eqn:CBF-QP} controller \cite{rosolia2020multi}. In practice, when the desired controller $\mb{k}_d$ is synthesized without considering safety, this can lead to aggressive behavior when the system approaches the boundary of the safe set. 
Using MPC in a multi-layered setup allows the safety constraint to be incorporated into the specification of $\mb{k}_d$. When the MPC directly operates on the full state and input of~\eqref{eq:full_rigid_body_dynamics}, the safety constraint in~\eqref{eqn:reldeg2} is readily incorporated, as was done in~\cite{zeng2020safety}. In contrast, we consider a MPC controller that operates on a reduced order model in which case the barrier constraint in~\eqref{eqn:reldeg1} is added to the MPC problem instead. For simplicity of exposition, we present here an MPC layer that operates on a purely kinematic model of the system.

Given a current estimate of the state $(\hat{\vq},\dot{\hat{\vq}})$ at time $\hat{t}$, a kinematic MPC solves the following optimization problem:


\boxedequation{\small Low-Frequency Safe Kinematic MPC}{
\small
\begin{align*}
\underset{\begin{subarray}{c}
		\vq^d(t), \dot{\vq}^d(t)
		\end{subarray}}{\min}\,\,\, &\Phi(\vq^d(T)) + \int_{0}^{T} L(\vq^d(t),\dot{\vq}^d(t), t) \, \dif t, 
\end{align*}
\vspace{-3mm}
\begin{align*}
	\text{s.t}\quad &\quad  \vq^d(0) = \hat{\vq}, \qquad \derp{\vq^d}{t} = \dot{\vq}^d, \\
	&\quad \dot{h}(\vq^d,\dot{\vq}^d,t)+\alpha_1(h(\vq^d,t)) \geq 0, 
\end{align*}
\vspace{-5mm}
}


\noindent where $\mb{q}^{d}(t)$ and $\dot{\vq}^{d}(t)$ are trajectories of generalized coordinates and velocities, forming the safe desired trajectory for the tracking controller. A desired acceleration is obtained through a combination of tracking terms and a forward difference of the desired velocities:
\begin{equation*}
    \ddot{\vq}^d = \frac{\dot{\vq}^d(\hat{t}+\delta t)-\dot{\vq}^d(\hat{t})}{\delta t} + \vD (\dot{\vq}^d(\hat{t}) - \dot{\hat{\vq}})  + \vP (\vq^d(\hat{t}) - \hat{\vq}). 
\end{equation*}



\noindent Drawing inspiration from the inverse dynamics approach in \cite{reher2020inverse}, the high-frequency controller is given by:

\boxedequation{\small High-Frequency ID-CBF-QP}{
\small
\begin{align}
\label{eqn:ID-CBF-QP}
k(\mb{q},\dot{\mb{q}},t) =  \,\underset{\bs{\tau},\,\ddot{\mb{q}}}{\argmin}&    \quad \frac{1}{2} \| \ddot{\mb{q}} -\ddot{\mb{q}}^d\|_2^2 \nonumber  \\
\mathrm{s.t.} \quad\quad  & \mb{D}(\mb{q})\ddot{\mb{q}}+\mb{C}(\mb{q},\dot{\mb{q}})\dot{\mb{q}}+\mb{G}(\mb{q}) = \mb{B}(\mb{q})\bs{\tau}, \nonumber\\ 
 & \dot{h}_e(\mb{q},\dot{\mb{q}},t,\bs{\tau})
 \geq -\alpha_2(h_e(\mb{q},\dot{\mb{q}},t)). \nonumber
\end{align}
\vspace{-5mm}
}
\noindent This controller seeks to track the desired acceleration determined by the low-frequency MPC controller while ensuring that the full dynamics are incorporated into the determination of safe inputs according to \eqref{eqn:reldeg2}.

\section{ANYmal Implementation}
\label{sec:implementation}
In this Section we provide an overview of how the multi-layer control formulation discussed in Section~\ref{sec:formulation} is applied to the ANYmal quadrupedal robotic platform. An overview of the control structure is provided in Figure \ref{fig:control_overview}. 


\subsection{MPC System Model}
We apply our approach to the kino-dynamic model of a quadruped robot, which describes the dynamics of a single free-floating body along with the kinematics for each leg. 
The state $\mb{x}\in\R^{24}$ and input $\mb{u}\in\R^{24}$ are defined as: 
\begin{equation}
\vx  = \begin{bmatrix} \vth^T, & \vp^T, & \vom^T, & \vv^T, & \vq^T \end{bmatrix}^T, \quad \vu  = \begin{bmatrix} \vlambda_{B}^T, & \dot{\mb{q}}^{d^T} \end{bmatrix}^T, \nonumber
\end{equation}
where $\vth\in\R^{3}$ is the orientation of the base in Euler angles, $\vp\in\R^3$ is the position of the center of mass in the world frame $\mathcal{F}_W$, $\vom\in\R^3$ is the angular rate of the base, $\vv\in\R^3$ is the linear velocity of the center of mass in the body frame $\mathcal{F}_B$, and $\vq\in\R^{12}$ is the joint positions. The joint positions for leg $i$ are given by $\mb{q}_i\in\R^3$. The inputs of the model are end-effector contact forces ${\vlambda_{B}}\in\R^{12}$ in the body frame and desired joint velocities $\dot{\mb{q}}^d\in\R^{12}$ with equations of motion:
\begin{align}
&
\begin{array}{ll}
		\dot{\vth} = \vT(\vth) \vom,  \qquad \dot{\vp}  = _W\!\vR_B(\vth) \, \vv,\\
		\dot{\vom} = \vI^{-1} \left(  -\vom \times \vI \vom + \sum_{i=1}^4{ {\vr_{B_i}}(\vq_i) \times {\vlambda_{B_i}}} \right), \\
		\dot{\vv}  = \vg(\vth) + \frac{1}{m} \sum_{i=1}^4{\vlambda_{B_i}}, \qquad
		\dot{\vq} = \dot{\mb{q}}^d,
\end{array}	 
 \notag
\end{align}
where $_W\!\vR_B:\R^3\to SO(3)$ is the rotation matrix from $\mathcal{F}_B$ to $\mathcal{F}_W$ and $\vT:\R^3\to\R^{3\times 3}$ transforms angular velocities to the Euler angles derivatives. Model parameters include the gravitational acceleration in the body frame $\vg:\R^3\to\R^3$, the total mass $m\in\R_+$, and the moment of inertia $\vI\in\R^{3\times3}$. The moment of inertia is assumed constant and taken at the upright state of the robot. We denote ${\vr_{B_i}}:\R^3\to\R^3$ as the position of foot $i$ relative to the center of the mass in the body frame.

\subsection{MPC Constraints}
In this subsection we list the constraints that are included in the low-frequency kino-dynamic MPC controller.

\subsubsection{Mode Constraints}
The mode constraints capture the different modes of each leg at any given point in time. We assume that the mode sequence is a predefined function of time. The resulting mode-dependent constraints are
\begin{align*}
&\left\{ 
\begin{array}{ll}
		\vv_{W_i}(\vx, \vu) = \mathbf{0}, \quad &\text{if $i$ is a stance leg}, \\
		\vn^T \vv_{W_i}(\vx, \vu)  = c(t), \quad  \vlambda_{B_i} = \mathbf{0}, \quad  &\text{if $i$ is a swing leg},
\end{array}
\right.
\end{align*}
where ${\vv_{W}}_i$ is the end-effector velocity in world frame.  
These constraints ensure that stance legs remain on the ground and a swing legs follow a predefined curve $c:\R_+\to\R$ in the direction of the local surface normal $\vn\in\R^3$ to avoid foot scuffing. 

\subsubsection{Friction Cone Constraints}
The end-effector forces are constrained to lie in the friction cone, $\vlambda_{W_i} \in \mathcal{Q}(\vn, \mu_c)$,
defined by the surface normal $\mb{n}$ and friction coefficient ${\mu_c = 0.7}$. After resolving the contact forces in the local frame of the surface, given by ${\vF = [F_x, F_y, F_z]}$, a second-order cone constraint is specified, $h_{cone} = \mu_c F_z - (F_x^2 + F_y^2)^{\frac{1}{2}} \geq 0$.

\begin{figure*}[b!]
    \centering
    \includegraphics[trim=0 40 0 40,clip,width=\linewidth]{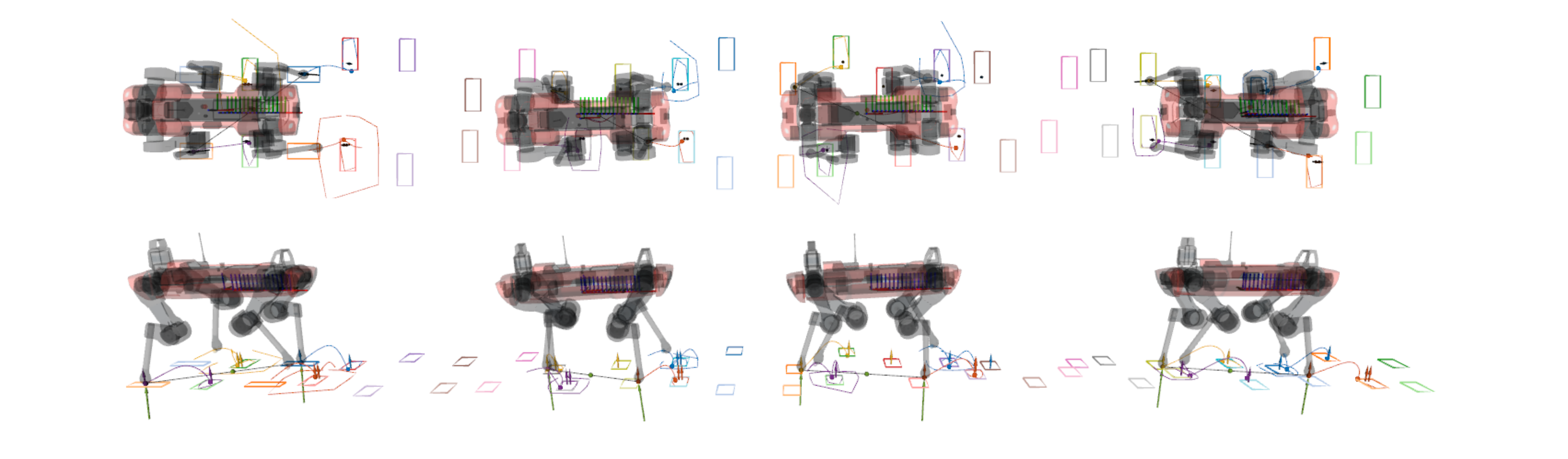}
    \caption{ANYmal traversing stepping-stones in simulation using the multi-layer CBF-MPC controller. The target foothold regions as well as the contracting barrier constraints are shown at snapshots in the motion. See the video in the supplementary material for the full motion~\cite{video}.}
    \label{fig:sim_forward_stones}
\end{figure*}

\subsubsection{State-Only Foot Placement Constraints}
\label{sect:state-only-foot-constraints}
When foot-placement is formulated as a state-only constraint (rather than encoded in a CBF), it is specified as the following inequality constraint on stance feet:
\begin{equation}
\new{\vh_i^{t}}(\vx) = \vA_i \cdot \vp_{W_i}(\vx) + \vb_i \geq \mathbf{0},
\label{eq:foothold_position_constraint}
\end{equation}
where $\mb{A}_i\in\R^{p_i\times 3}$, $\mb{b}_i\in\R^{p_i}$, and $\vp_{W_i}:\R^{24}\to\R^3$ is the position of foot $i$ in the world frame. The matrix $\mb{A}_i$ and $\mb{b}_i$ project the position of foot $i$ on to the target terrain and form a set of half-space constraints to ensure the foot lands within a desired target region. Instead of constraining the stance feet, a similar constraint can be placed on the swing feet with a constraint set that shrinks in time and converges to the desired foot placement region:
\begin{equation}
\new{\vh_{i}^{w}}(\vx, t) = \vA_i \cdot \vp_{W_i}(\vx) + \vb_i + s(t) \cdot \mb{1} \geq \mb{0},
\label{eq:foothold_position_barrier_constraint}
\end{equation}
where $s:\R_+\to\R_+$ converges to $0$ as the $t$ approaches the duration of the swing phase.

\subsubsection{Barrier Foot Placement Constraints}
\label{sect:barrier-foot-constraints}
When posed as a CBF constraint as in the proposed low-frequency Safe Kinematic MPC controller, the foot placement constraints are specified with constant $\gamma \in\R_{++}$ as:
\begin{equation}
\new{\vh_{e,i}^{w}}(\vx, \dot{\vq}, t) = \new{\dot{\vh}_{i}^{w}}(\vx, t, \vu) + \gamma \new{\vh_{i}^{w}}(\vx, t) \geq \mathbf{0}.
\label{eq:foothold_mpc_barrier_constraint}
\end{equation}

\subsection{Whole-Body Tracking Control}
\label{sect:Whole-Body-Tracking-Control}
The control signal $\vu$ determined by the low-frequency MPC layer consists of contact forces and desired joint velocities. A high-frequency hierarchical inverse dynamics controller is used to convert the optimized MPC trajectory into torque commands \cite{bellicoso2016perception}. This whole body control (WBC) approach considers the full nonlinear rigid body dynamics of the system. At each priority, a QP is solved in the null space of all higher priority tasks. Each task is a equality or inequality constraint that is affine in the generalized accelerations, torques, and contact forces. The CBF constraints, which are by design affine in the control torques, are therefore readily integrated into this framework. The full list of tasks is given in Table~\ref{tab:controllertasks}. 

As described in Section \ref{sec:formulation}, the following CBF constraint incorporating the dynamics can be included in the whole-body controller:
\begin{equation}
\new{\dot{\vh}_{e,i}^{w}}(\vx, \dot{\vq},t, \vtau) + \xi \new{\vh_{e,i}^{w}}(\vx, \dot{\vq}, t) \geq \mathbf{0}
\label{eq:foothold_wbc_barrier_constraint}
\end{equation}
with $\xi\in\R_{++}$. Finally, the torque derived from the whole body controller, $\bs{\tau}_{WBC}\in\R^{12}$, is computed. To compensate for model uncertainty for swing legs \new{(on hardware, not in simulation)}, the integral of joint acceleration error with gain $K\in\R_{++}$ is added to the torque applied to the system: 
\begin{equation}
\vtau = \vtau_{WBC} - K \int_{t^{sw}_{0}}^t \left( \ddot{\vq} - \ddot{\vq}_{WBC}  \right) \text{d}t
\end{equation}
\new{While this modification implies $\bs{\tau}$ may not satisfy the CBF condition in \eqref{eq:foothold_wbc_barrier_constraint}, we note that $\bs{\tau}_{WBC}$ may not satisfy \eqref{eq:foothold_wbc_barrier_constraint} in the presence of model uncertainty. To achieve safe behavior in practice, it is necessary to balance the choice of safe inputs with model uncertainty.}

\begin{table}[bt]
\centering
\caption{WHOLE-BODY CONTROL TASK HIERARCHY.}
\footnotesize
\begin{tabular}{|c|c|l|}
\hline
Priority & Type &Task \\
\hline
0 & $=$ & Floating base equations of motion. \\
  & $\geq$ & Torque limits. \\
  & $\geq$ & Friction cone constraint. \\
  & $=$ & No motion at the contact points. \\
  & $\geq$ & \textbf{Control barrier constraints.} \\
\hline
1 & $=$ & Torso linear and angular acceleration. \\
  & $=$ & Swing leg motion tracking. \\
\hline  
2 & $=$ & Contact force tracking. \\
\hline
\end{tabular}
\label{tab:controllertasks}
\end{table}

\subsection{User Commands \& Terrain Selection}
User commanded twists and a desired gait pattern are provided to the robot via joystick and extrapolated to a state reference signal $\vx_{ref}(t)$. The reference input $\vu_{ref}(t)$ is constructed by equally distributing the weight over all contact feet. The MPC cost function is a frequency dependent quadratic cost around the reference trajectories to promote smooth optimal inputs \cite{grandia2018frequency}.

We assume that a segmented terrain model with each segment described by a planar boundary and a surface normal is available. For each contact phase within the MPC horizon, the terrain segment is selected that is closest to the reference end-effector position determined by $\vx_{ref}(t)$, evaluated at the middle of the stance phase. A convex polygon is fit to the selected terrain, starting from the reference end-effector position projected onto the segment boundary. This polygon, together with the surface normal, define the half spaces for the constraints in \eqref{eq:foothold_position_constraint} and \eqref{eq:foothold_position_barrier_constraint}.

\section{Results}
\label{sec:simulation}

We evaluate the controller proposed in Section \ref{sec:implementation} in simulation on a classical stepping-stones scenario as shown in Figure \ref{fig:sim_forward_stones}. The stones are configured with a pattern of \SI{0.5}{\meter} width and \SI{0.35}{\meter} longitudinal spacing, with random displacements up to 10, 15, and \SI{5}{\cm}, in longitudinal, lateral, and vertical direction respectively. The controller is commanded to perform a trotting gait with a forward velocity of \SI{0.25}{\meter/\second}, and commanded to stop on the final stone. We compare our proposed controller, \new{numbered V,} against \new{four} alternative formulations and report results in Table~\ref{tab:sim_results}.

\begin{table}[h]
\centering
\caption{SIMULATION RESULTS}
\footnotesize
\setlength\tabcolsep{4.0pt}
\begin{tabular}{ 
>{\arraybackslash}m{2.4cm}
>{\centering\arraybackslash}m{0.9cm}
>{\centering\arraybackslash}m{0.9cm}
>{\centering\arraybackslash}m{0.9cm}
>{\centering\arraybackslash}m{0.9cm}
>{\centering\arraybackslash}m{0.9cm}}
\hline
                    
                             & I &  II & III & \new{IV} & V \\ \hline
\new{MPC constr.}                 & \new{None}    & \new{CBF}        & \new{State}        &  \new{State}  & \new{CBF}          \\
\new{WBC constr.}          & \new{CBF}    & \new{None}        & \new{None}         &  \new{CBF}    & \new{CBF}          \\ \hline
num. steps                         & 28                 & 140        & 140    & \new{140}  & 140  \\
num. missteps                      & 5                  & 6          & 5      & \new{0} & \new{0}    \\
avg. misstep {[}mm{]}              & 1.4                & 2.5        & 4.3    & \new{-} & \new{-} \\ \hline
total swing time {[}s{]}           & 11.0               & 49.0       & 48.6   & \new{48.4} & \new{48.6} \\
$\new{\vh^{w}_{i}}<0$ time {[}s{]}      & 2.4                & 2.3        & 15.3 & \new{2.6} & \new{0.4} \\
$\new{\vh^{w}_{e,i}}<0$ time {[}s{]}    & 3.3                & 5.4        & 15.6 & \new{3.7} & 0.8 \\ \hline
\end{tabular}
\label{tab:sim_results}
\end{table}

As seen in the supplementary video~\cite{video}, the controller with no foot placement constraints in the MPC controller and a CBF constraint in the high-frequency controller (denoted CBF-QP, and the closest to the related work \cite{nguyen20163d}) is able to enforce safety for a number of steps, but quickly destabilizes. The absence of information on the safety constraint in the MPC layer results in an abrupt and strong correction for safety by the high-frequency CBF. This approach work well only when the stepping-stones are placed close to the nominal gait of the robot, but it fails in this more challenging scenario.

The second and third controllers include foot placement constraints in the MPC controller, but not in the high-frequency controller. In the second controller the constraints are implemented as CBFs \new{through \eqref{eq:foothold_mpc_barrier_constraint}} and in the third controller they are implemented as state constraints \new{through \eqref{eq:foothold_position_constraint}}. Both of these controllers are able to successfully traverse the length of the stepping-stones scenario. We see that the controllers exhibit similar numbers of missteps, but the MPC controller with CBFs has smaller average misstep size. 

\begin{figure}[tb]
    \centering
    \begin{minipage}[c]{0.60\columnwidth}
        \centering
        \includegraphics[trim=35 0 110 335,clip,width=\columnwidth]{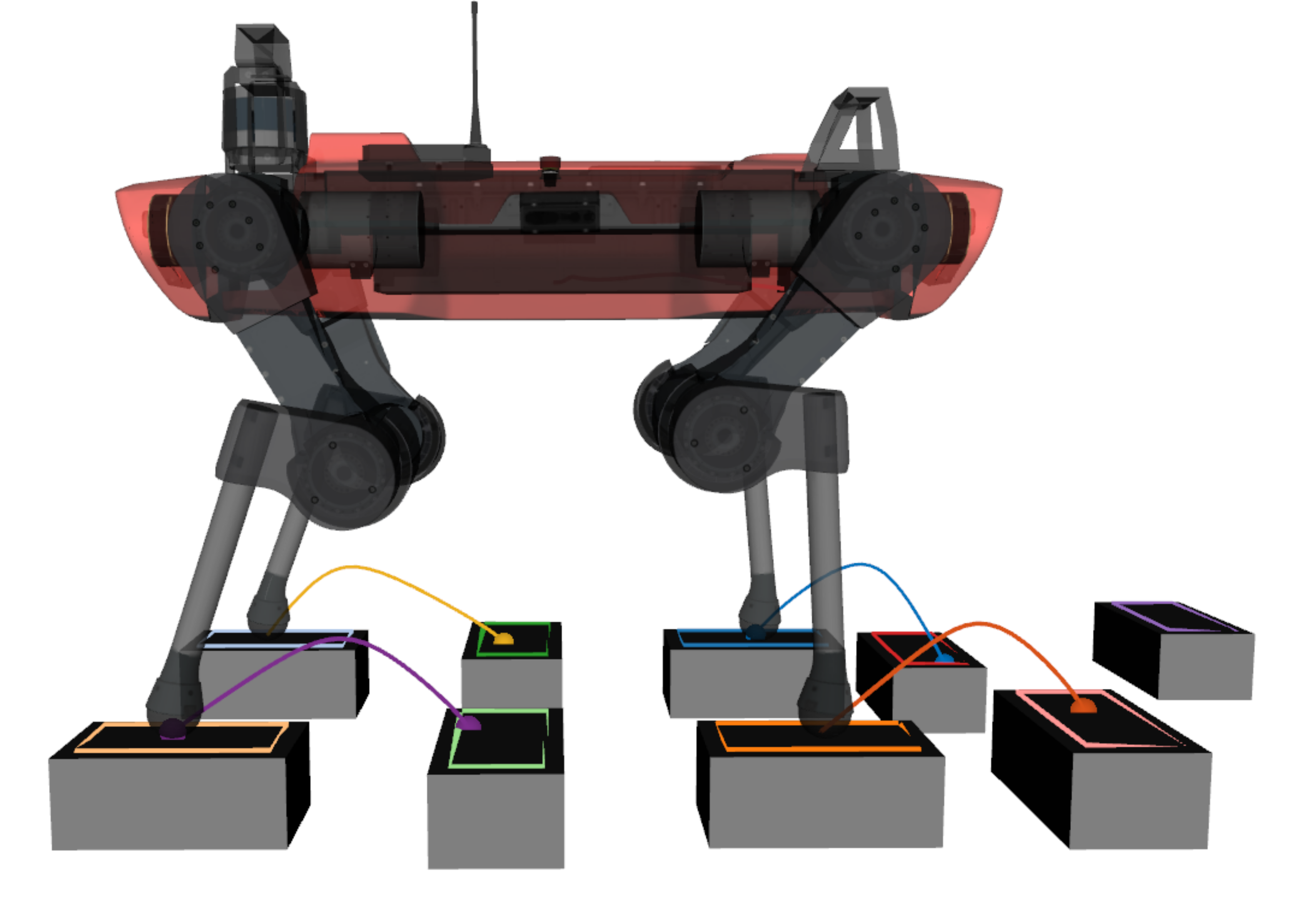}
    \end{minipage}
    \hfill
    \begin{minipage}[c]{0.35\columnwidth}
        \centering
        \includegraphics[width=\columnwidth]{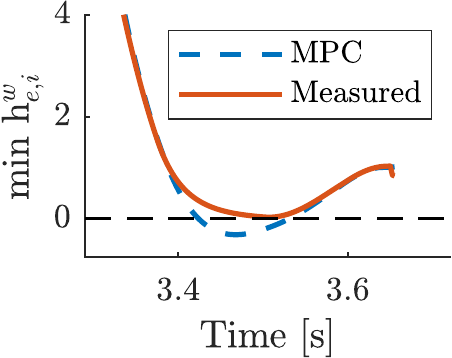}
    \end{minipage}
    \begin{minipage}{0.60\columnwidth}
        \centering
        \includegraphics[trim=60 0 90 335,clip,width=\columnwidth]{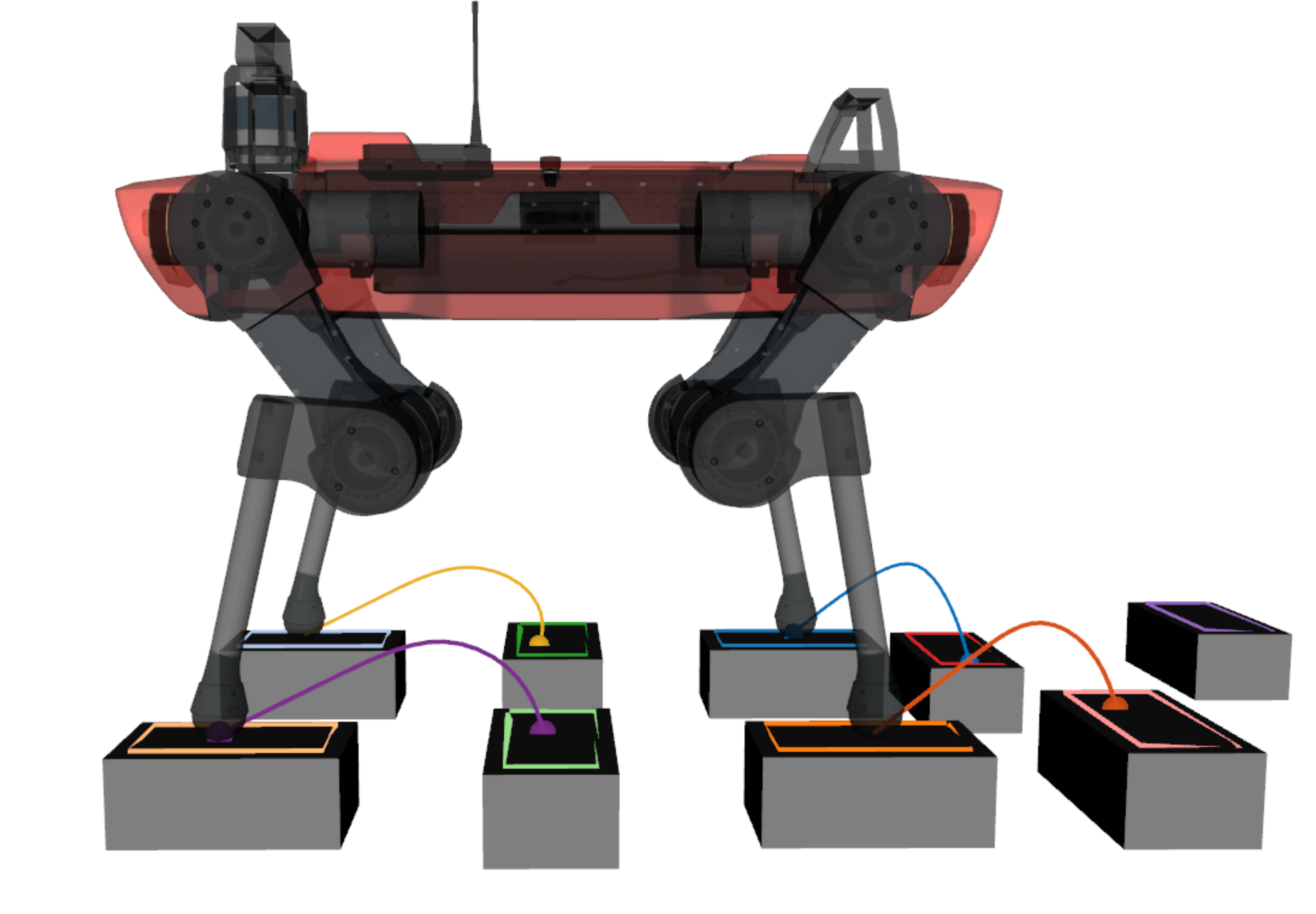}
    \end{minipage}
    \hfill
    \begin{minipage}{0.35\columnwidth}
        \centering
        \includegraphics[width=\columnwidth]{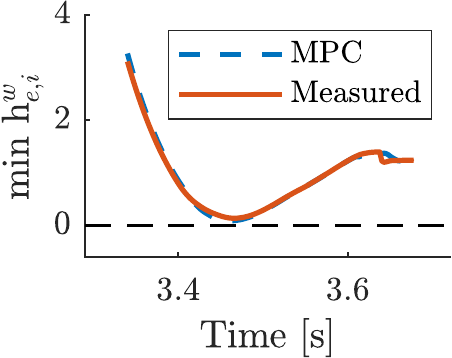}
    \end{minipage}
    \caption{\new{Visualization of the planned MPC trajectories for different constraint formulations. \textbf{Top:} MPC with state constraints on the touchdown location (controller IV). \textbf{Bottom:} MPC with CBF constraints (controller V). The plots on the right show the planned and measured values of $\mb{h}^{w}_{e,i}$ for the right front foot, with deviations from the MPC optimal trajectory occurring when CBF constraints are absent from the MPC formulation.}}
    \vspace{-0mm}
    \label{fig:swing_comparison}
\end{figure}

\new{The fourth and fifth controller enforce the high frequency CBFs \eqref{eq:foothold_wbc_barrier_constraint} and contain either state constraints \eqref{eq:foothold_position_constraint} or CBFs \eqref{eq:foothold_mpc_barrier_constraint} in the MPC formulation. Both controllers complete the scenario without missteps. However, the proposed controller shows the least amount of time violating the barrier conditions. The difference can be explained through the results in Figure~\ref{fig:swing_comparison}. Because the MPC with state constraints (top) is not aware of the CBF condition, it plans for a trajectory that violates these constraints during the swing phase. During execution, the high-frequency tracking controller strictly enforces the CBF, resulting in a deviation from the MPC plan. Such abrupt deviations can cause problems, for example when operating close to kinematic limits. Consistently enforcing the CBF condition removes this mismatch (bottom).} 

\begin{figure}[t]
    \centering
    \begin{minipage}{\linewidth}
        \centering
        \includegraphics[trim=0 10 0 0, clip, width=0.95\linewidth]{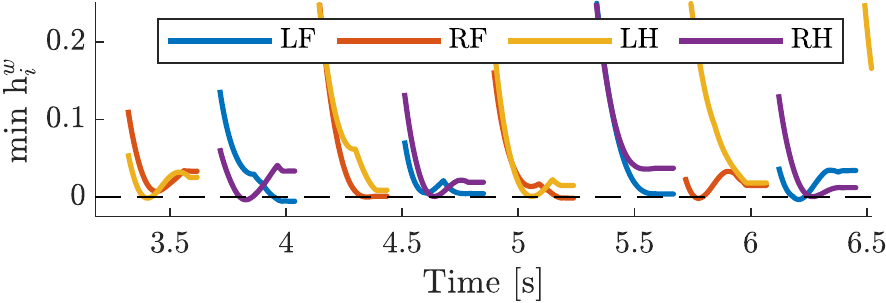}
    \end{minipage}
    \hfill
    \begin{minipage}{\linewidth}
        \centering
        \includegraphics[width=0.9\linewidth]{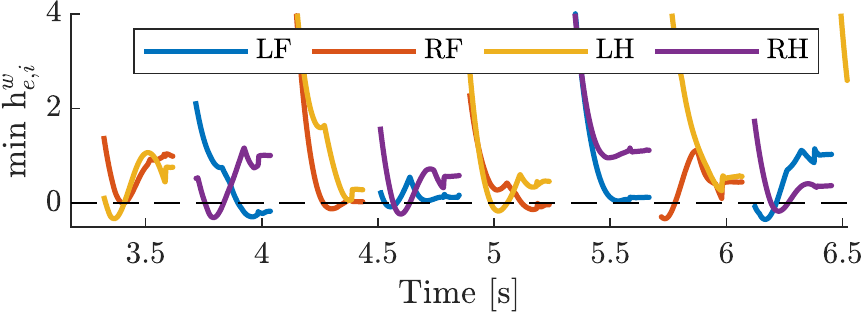}
    \end{minipage}
    \caption{The minimum values of $\new{\mb{h}^{w}_i}$ and $\new{\mb{h}^{w}_{e,i}}$ per leg during the stepping-stones simulation \new{with the CBF only at MPC level}.}
    \label{fig:hvalue_sim_CBF_MPCOnly}
    \vspace{-4mm}
\end{figure}

\begin{figure}[t]
    \centering
        \begin{minipage}{\columnwidth}
        \centering
        \includegraphics[trim=0 10 0 0, clip,width=0.95\columnwidth]{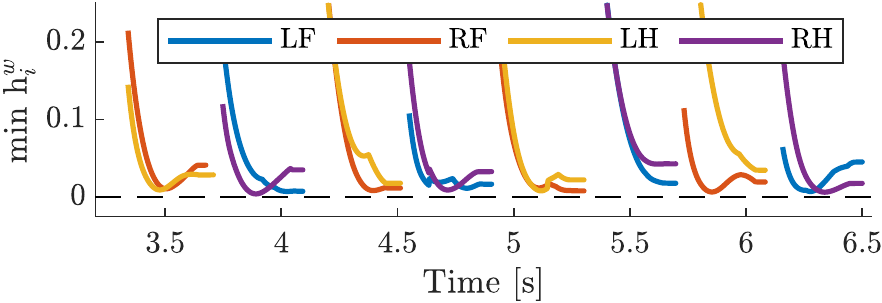}
    \end{minipage}
    \hfill
    \begin{minipage}{\columnwidth}
        \centering
        \includegraphics[width=0.9\columnwidth]{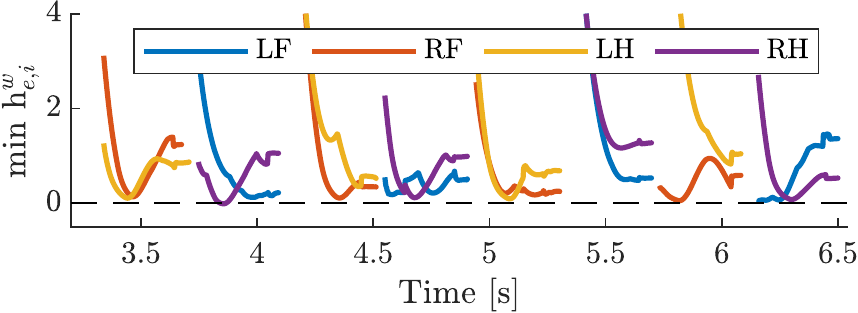}
    \end{minipage}
    \caption{The minimum values of $\new{\mb{h}^{w}_i}$ and $\new{\mb{h}^{w}_{e,i}}$ per leg during the stepping-stones simulation \new{with the CBF in both WBC \& MPC}.}
    \label{fig:hvalue_sim_CBF_Both}
\end{figure}

\new{The simulation experiments indicate that including CBF constraints in the high-frequency controller leads to safer behavior, and that including terrain constraints in the MPC controller prevents the high-frequency CBF from destabilizing the gait. Finally, enforcing CBF constraints in both layers of the hierarchy prevents an inconsistency that results in large deviations from the optimal solution determined by the MPC layer.} The values of $\new{\mb{h}^{w}_i}$ and $\new{\mb{h}_{e,i}^{w}}$ for the controller with CBFs only in the MPC and for with the CBFs in both WBC \& MPC can be seen in Figures \ref{fig:hvalue_sim_CBF_MPCOnly} and \ref{fig:hvalue_sim_CBF_Both}. The controller with CBFs at both levels has smaller violations of constraints \eqref{eq:foothold_position_barrier_constraint} and \eqref{eq:foothold_mpc_barrier_constraint}.



We evaluate the efficacy of this method experimentally on the ANYmal robotic platform. All computation runs on a single onboard PC (Intel i7-8850H, \SI{2.6}{\GHz}, hexa-core 64-bit) with the MPC solver running asynchronously at \SI{30}{\hertz} and the whole-body QP tracking controller running at \SI{400}{\hertz}.

\begin{figure}[t!]
    \centering
        \begin{minipage}{\columnwidth}
        \centering
        \includegraphics[trim=0 10 0 0, clip,width=0.95\columnwidth]{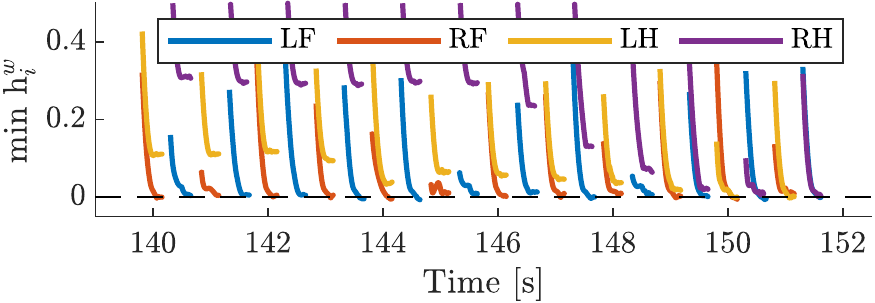}
    \end{minipage}
    \hfill
    \begin{minipage}{\columnwidth}
        \centering
        \includegraphics[width=0.93\columnwidth]{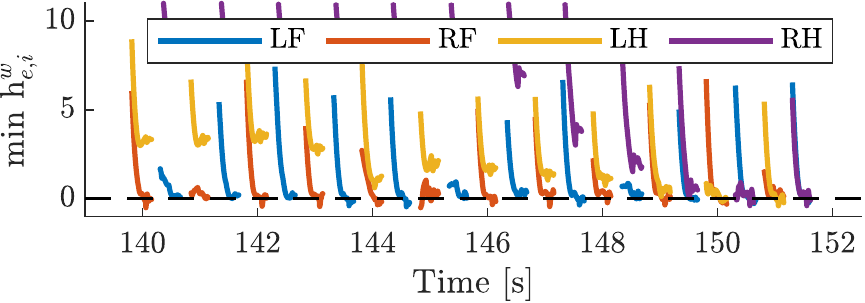}
    \end{minipage}
    \caption{The minimum values of $\new{\mb{h}^{w}_i}$ and $\new{\mb{h}^{w}_{e,i}}$ per leg during the stepping-stones \new{hardware} experiment for \new{the proposed controller with CBF constraints in WBC \& MPC}.}
    \vspace{-3mm}
    \label{fig:hvalue_hardware}
\end{figure}

The robot is initialized on pre-mapped terrain and receives external base twist and gait commands. The size of the segmented regions are decreased by \SI{5}{\cm} with respect to the real boundary to provide a margin for state estimation errors.  In the supplementary video~\cite{video} we visualize the internal state of the controller. For legs that are in swing, a projection of the barrier constraint in~\eqref{eq:foothold_position_barrier_constraint} onto the terrain is plotted. This barrier constraint shrinks over time and converges to the selected target foothold region at foot contact. Furthermore, it can be seen how the foothold target is large when stepping onto the wooden pallet. This shows that the proposed method can seamlessly transition between rough and flat terrain, restricting the motion only when necessary for safe foot placement.  The values of $\new{\mb{h}^{w}_i}$ and $\new{\mb{h}_{e,i}^{w}}$ for several steps can be seen in Figure~\ref{fig:hvalue_hardware}. Both constraints are rarely violated, which confirms that the safety constraints are successfully transferred to hardware.

\section{Conclusions}
\label{sec:conclusions}
We proposed a multi-layered control framework that combines CBFs with MPC. Simulation experiments show that enforcing CBF constraints on both the MPC and QP tracking layer outperforms variants where they are enforced at only one of the layers. Additionally, we validated the viability of the approach on hardware by demonstrating dynamic locomotion on stepping-stones with safety constraints. Future work includes developing a perception pipeline to automatically perform terrain-based segmentation from sensor data and studying the theoretical properties of the proposed controller.


\clearpage
\bibliographystyle{bibtex/myIEEEtran} 
\bibliography{bibtex/IEEEabrv,bibtex/taylor_main,bibtex/references}


\end{document}